# Fine-grained Multi-class Nuclei Segmentation with Molecular-empowered All-in-SAM Model

**Xueyuan Li**[a], **Can Cui**[b], **Ruining Deng**[b], **Yucheng Tang**[d], **Quan Liu**[b], **Tianyuan Yao**[b], **Shunxing Bao**[c], **Naweed Chowdhury**[e], **Haichun Yang**[f], **Yuankai Huo**[a,b,c,f,*]

[a]Data Science Institute, Vanderbilt University, Nashville, TN, USA
[b]Department of Computer Science, Vanderbilt University, Nashville, TN, USA
[c]Department of Electrical and Computer Engineering, Vanderbilt University Medical Center, Nashville, TN, USA
[d]NVIDIA Corporation, Redmond, WA, USA
[e]Department of Otolaryngology–Head and Neck Surgery, Vanderbilt University Medical Center, Nashville, TN, 37232, USA
[f]Department of Pathology, Microbiology and Immunology, Vanderbilt University Medical Center, Nashville, TN, 37232, USA

**Abstract.** **Purpose:** Recent developments in computational pathology have been driven by advances in Vision Foundation Models, particularly the Segment Anything Model (SAM). This model facilitates nuclei segmentation through two primary methods: prompt-based zero-shot segmentation and the use of cell-specific SAM models for direct segmentation. These approaches enable effective segmentation across a range of nuclei and cells. However, general vision foundation models often face challenges with fine-grained semantic segmentation, such as identifying specific nuclei subtypes or particular cells. **Approach:** In this paper, we propose the molecular-empowered All-in-SAM Model to advance computational pathology by leveraging the capabilities of vision foundation models. This model incorporates a full-stack approach, focusing on: (1) annotation—engaging lay annotators through molecular-empowered learning to reduce the need for detailed pixel-level annotations, (2) learning—adapting the SAM model to emphasize specific semantics, which utilizes its strong generalizability with SAM adapter, and (3) refinement—enhancing segmentation accuracy by integrating Molecular-Oriented Corrective Learning (MOCL). **Results:** Experimental results from both in-house and public datasets show that the All-in-SAM model significantly improves cell classification performance, even when faced with varying annotation quality. **Conclusions:** Our approach not only reduces the workload for annotators but also extends the accessibility of precise biomedical image analysis to resource-limited settings, thereby advancing medical diagnostics and automating pathology image analysis.

*****Corresponding Author:** Yuankai Huo, yuankai.huo@vanderbilt.edu

## 1 Introduction

The field of computational pathology,[1] is undergoing transformative advancements by integrating computational algorithms with whole slide imaging (WSI). This integration has shown promising results with improved diagnostic precision and advancing personalized medical treatments.[2] Computational pathology focuses on analyzing digital pathological images to support clinical decisions and personalized therapies.[3] However, accurately segmenting and classifying cell nuclei in these images remains challenging, especially in oncology, where it significantly impacts diagnostic and therapeutic planning.

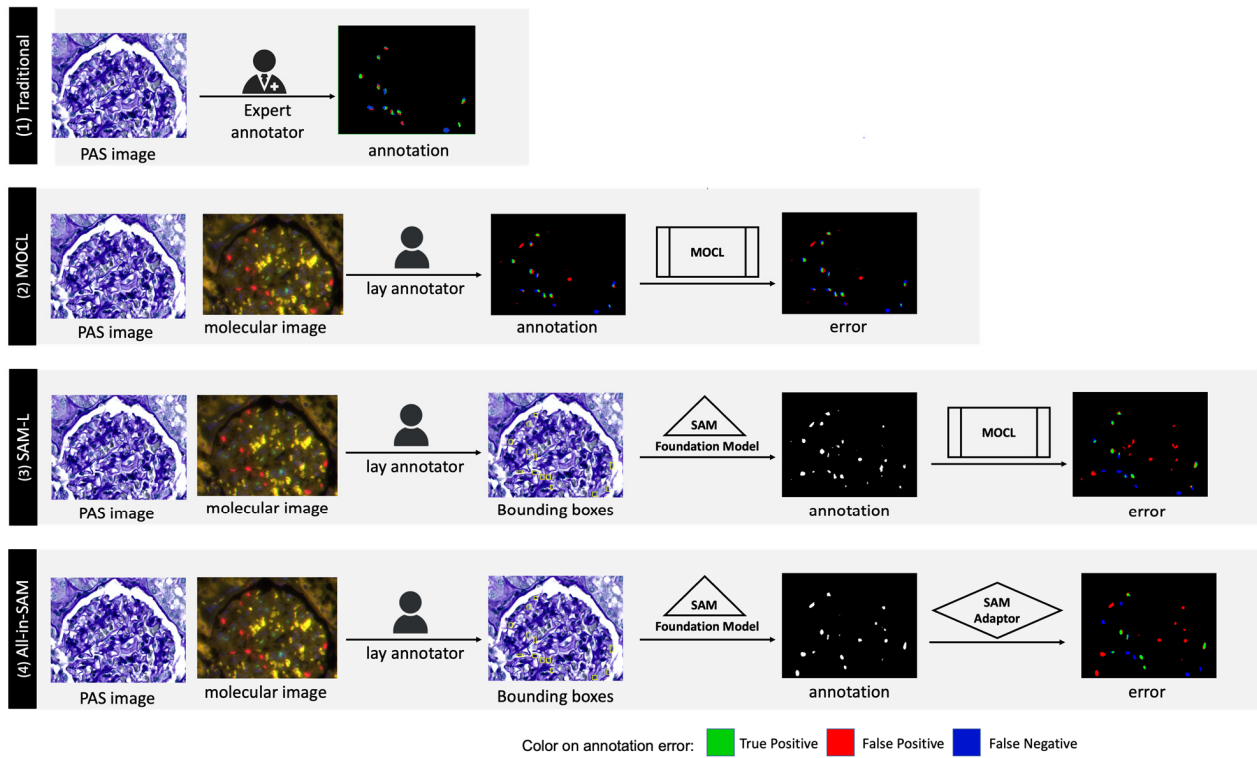

Fig 1: **Overall Idea of our work**: This diagram illustrates the distinctions between our approach (bottom panel) and existing methods. (1) Traditional: Expert annotators manually label cells using only PAS images. (2) MOCL: Lay annotators provide pixel-level labels under the guidance of immunofluorescence (IF) molecular images, followed by the application of deep learning for segmentation. (3) SAM-L: The SAM technique is utilized to expedite the annotation process, requiring only minimal (box) annotations. (4) All-in-SAM (Our Method): We integrate SAM in the annotation phase and adaptively fine-tune it during the training of the model.

Recently, Vision Foundation Models (VFMs)[4,5,6] have made significant strides in the area of medical image segmentation. These models are typically trained on large and diverse datasets, to achieve superior performance across different tasks. The Segment Anything Model (SAM),[5,7,8] as an example, has been particularly notable for its adaptability and efficiency, offering effective segmentation across various scenarios with minimal need for detailed annotations.[9,10] In digital pathology, however, VFMs face challenges in performing fine-grained semantic segmentation that is necessary for distinguishing subtle differences among cell types or accurately segmenting specific cells in heterogeneous tissues.[11, 12] This limitation is critical in clinical studies and scientific research.

In this paper, we propose the molecular-empowered All-in-SAM model, a holistic framework designed for precise cell segmentation and nuclei classification (Fig. 1). Our model adopts a full-stack approach, which comprises: (1) annotation—employing molecular-empowered learning to engage lay annotators and minimize the need for intricate pixel-level annotations, (2) learning—modifying the SAM model to focus on specific semantics, thereby capitalizing on its robust generalizability through the use of an SAM adapter, and (3) refinement—boosting segmentation accuracy by incorporating Molecular-Oriented Corrective Learning (MOCL).

In the proposed All-in-SAM framework, the VFM leverages both annotation generation and model fine-tuning stages, to reduce annotation costs while maintaining high segmentation performance for instance nuclei segmentation. To address the key challenge of fine-grained instance cell segmentation for different cell types, our MOCL approach incorporates molecular imaging data in training (but not needed in inference), multi-modal registration, and corrective learning, providing a holistic solution for this challenging task.

The contribution of the proposed All-in-SAM model is four-fold:

- *Scheme*. **Molecular-empowered learning allows fine-grained nuclei annotation**: We present molecular-empowered learning, a method that reduces the need for extensive domain expertise in detailed annotation. This approach allows non-experts, such as undergraduate students, to accurately annotate fine-grained nuclei from histopathology data using minimal domain knowledge. This is achieved by incorporating paired molecular immunofluorescence

(IF) images, lowering the training costs associated with procuring annotated datasets from domain experts.

- *Annotation*. **SAM for annotation**: The SAM model is used to reduce the annotation workload by shifting from detailed pixel-level contour delineation to more efficient weak annotations, such as box annotations.

- *Learning*. **SAM adaptor for label-efficient finetuning**: The incorporation of the SAM adaptor enables the model to efficiently adapt pre-trained SAM model using the mentioned annotations, reducing the necessity for extensive retraining with newly labeled data.

- *Refinement*. **Advanced corrective learning techniques for segmentation refinement**: MOCL is introduced to further refine segmentation by correcting errors using molecular insights and partially annotated data, enabling more precise identification and segmentation of cell types, especially in complex and heterogeneous samples.

The proposed All-in-SAM model has been tested on both public and in-house dataset. The results show its superior performance on multi-class instance segmentation. It offers a promising direction for deploying pathological nuclei segmentation and classification leveraging the latest advancements in VFM.

## 2 Related Work

The convergence of molecular biology and digital imaging within the realm of biomedical analysis has led to significant strides in disease diagnosis and research. This section delves into pivotal advancements in nuclei identification and segmentation, as well as the evolution of foundation models that form the scaffolding for molecular analysis, framing the backdrop for the novel contributions of our study.

## 2.1 Cell and Nuclei Segmentation

Digital pathology has made significant strides in cell and nuclei segmentation, which is vital for detailed pathological analysis. The Glo-In-One toolkit[13] simplifies the detection of glomerular structures, integrating complex tasks into a user-friendly interface. Similarly, Juang et al.[14] have enhanced cellular segmentation within renal pathology by combining deep learning with generative morphology techniques.

Advancements in self-supervised learning have revolutionized nuclei segmentation. Xie et al.[15] introduced a model that uses data's intrinsic properties to facilitate training without extensive labeled datasets, improving both automation and accuracy. Additionally, Sahasrabudhe et al.[16] have implemented attention mechanisms within their self-supervised learning framework to significantly enhance nuclei segmentation, adapting well to the variability of nuclear morphology.

These innovations are crucial for accurate diagnosis in medical imaging, as seen in Kumar et al. ,[17] who provided essential resources for reliable nuclear segmentation. Despite these advancements, challenges persist due to the variability in morphology across different slides and the vast amount of data in each image.[18, 19]

Overall, the continuous evolution of segmentation technologies promises to refine the precision and utility of digital pathology, driving forward more advanced and automated methods for handling complex pathological data.

*2.2 Vision Foundation Models*

Vision Foundation Models (VFMs) have revolutionized many areas within computer vision due to their ability to generalize effectively from vast, diverse datasets to specific tasks with minimal additional training.[5, 6] These models, extensively pre-trained on large image datasets, are exceptionally adaptable and capable of managing complex visual tasks, which makes them indispensable in fields ranging from general computer vision to specialized medical imaging applications.

In medical imaging and pathology, VFMs such as Convolutional Neural Networks (CNNs), Vision Transformers (ViTs), and more recently, models like Swin Transformers and Perceiver IO, have shown significant promise. Each of these models brings unique strengths to the challenges inherent in pathological image analysis.

- **Convolutional Neural Networks (CNNs)**: CNNs have been foundational in image analysis, renowned for their efficiency in processing pixel data and extracting hierarchical features. In pathology, they are predominantly utilized for tasks such as tumor detection and tissue classification, showcasing their adeptness in handling detailed and nuanced analyses.[20, 21]

- **Vision Transformers (ViTs)**: Adapted from the transformer architecture initially developed for natural language processing, ViTs apply self-attention mechanisms to capture global dependencies within images. This characteristic enables them to excel in identifying intricate patterns and anomalies in medical images, which is crucial for complex diagnostic tasks.[22, 23]

- **Advanced Models (Swin Transformers and Perceiver IO)**: These newer VFMs enhance the capacity to manage diverse visual representations and complexities, making them highly suitable for the multifaceted nature of pathological image analysis.

- **Segment Anything Model (SAM)**: SAM is a groundbreaking advancement in VFMs, designed to offer highly flexible and generalizable image segmentation capabilities. Its proficiency in segmenting diverse and intricate objects within images is particularly beneficial in pathology, where distinguishing between normal and abnormal tissues is often challenging.

The integration of VFMs into pathology workflows significantly enhances diagnostic accuracy, speeds up decision-making, and automates routine analysis tasks, thereby reducing the workload for medical professionals and improving the scalability of assessments. As these models continue to evolve, their adoption in clinical pathology is poised to expand, promising to transform medical diagnostics and enable the development of more sophisticated, personalized treatments. The future of VFMs in pathology is bright, with ongoing technological advancements likely to introduce new capabilities and reshape the landscape of the field.

2.3 Foundation Models for Nuclei Segmentation

Advancements in deep learning, particularly through the development of foundation models like the Segment Anything Model (SAM) and UNet, have significantly enhanced the segmentation of nuclei in pathology. SAM, developed by Deng et al., showcases a pivotal shift towards adaptable, scalable models with zero-shot learning capabilities, reducing the dependency on heavily annotated datasets. This allows for effective generalization across various nuclei segmentation tasks without extensive dataset-specific training.[24]

Further advancements by Kaur et al. have refined the UNet architecture to autonomously segment nuclei in whole slide images (WSIs), highlighting how automation in deep learning can streamline the detection and analysis of key structures in medical imaging.[25] These automated methods are crucial for improving the throughput and diagnostic accuracy in pathological analysis.

Additionally, the integration of molecular data has been instrumental in refining the segmentation accuracy. Techniques like Cellpose 2.0 [9] and Stardist ,[10] introduced by Pachitariu and Stevens, respectively, enhance nuclei detection by providing high-precision segmentation in complex tissue, demonstrating significant improvements in tasks such as circulating tumor cell detection.

The Segment Any Cell framework by Na and the CellVit model by Horst further exemplify the evolution of foundation models. These models use advanced machine learning techniques to segment and classify different types of cells with remarkable accuracy, even under challenging conditions.[11, 12] This highlights the potential of foundation models not just in enhancing existing applications but also in pioneering new methods for medical diagnostics.

These technological innovations have set a new standard in the field, combining the robustness of foundation models with the precision required for effective nuclei segmentation. The synergy between these models and molecular-oriented corrective learning (MOCL) addresses key challenges in the field, paving the way for groundbreaking advancements in biomedical image analysis.

## 3 Method

Our proposed framework is illustrated in Fig. 2. Step 1 transforms weak annotations into pixel-level annotations using both PAS and IF images. SAM allow annotators to only provide weak annotation (e.g., box) by converting those to pixel-level annotation. Then, SAM adapter is introduced to finetune SAM model for multi-class nuclei segmentation. Last, the corrective learning is developed to further enhance the segmentation performance.

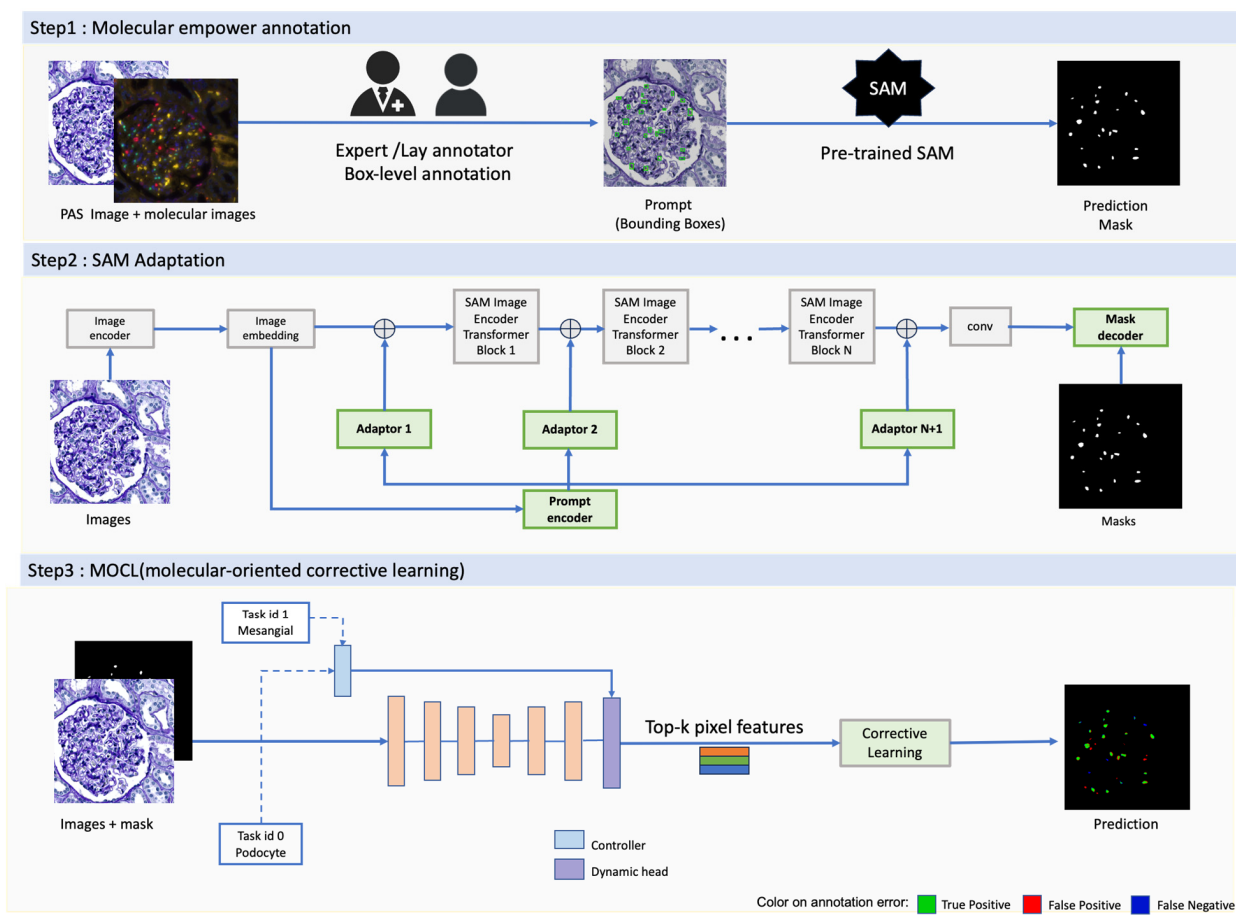

Fig 2: **Framework**: This figure shows the framework of the proposed All-in-SAM method. The framework consists of three main steps: (1) Molecular empower annotation, where experts or lay annotators provide box-level annotations to PAS images, which serve as prompts for the pre-trained SAM; (2) SAM Adaptation, which utilizes image embeddings processed through multiple transformer blocks and adaptors, integrating prompts to fine-tune the segmentation masks; and (3) MOCL (Molecular-Oriented Corrective Learning), where corrective learning processes are applied to refine segmentation based on top-k pixel features from images, enhancing the specificity and accuracy of the predictions.

## 3.1 Molecular-empowered SAM Annotation

Following our recent work,[26] the first major innovation of our pipeline is the integration of molecular-empowered learning into the annotation process. Specifically, we provide Periodic Acid-Schiff (PAS) and immunofluorescence (IF) images simultaneously, enabling lay annotators to perform fine-grained, multi-class instance segmentation of cell nuclei. This approach simplifies the segmentation task from identifying complex morphologies to merely recognizing different colors. Subsequently, all cell types are categorized using multiple binary masks.

In our All-in-SAM framework, SAM is employed to further simplify and accelerate the above molecular-empowered annotation process. Specifically, the lay annotators only draw the bounding boxes for cell nuclei. Then, such weak annotations are converted to pixel-level annotation using the SAM model.

## 3.2 SAM Adapter for Cell Segmentation

While SAM demonstrates strong performance in generic segmentation tasks, its efficacy may diminish when applied to specific tasks, potentially resulting in suboptimal outcomes or failures. To address this challenge, we utilize a new pipeline, presented in Figure 2 at step 2. This approach is discerning, opting not for a wholesale fine-tuning of the entire model but rather for a targeted adaptation of its latter layers.

More specifically, this strategy involves automatically extracting and encoding texture information from each image as handcrafted features, which are subsequently incorporated into multiple layers of the encoder. Supervised by approximate segmentation masks, this prompt-based fine-tuning is applied to the pretrained SAM model. During inference, nuclei can be segmented directly from images without the need for box prompts.

### 3.3 MOCL-assisted Segmentation Refinement

MOCL aims to further improve the segmentation performance of the SAM adapter as a post-processing step. Specifically, as illustrated in Figure 2 at step 3, MOCL utilizes top-k pixel feature embeddings from the annotation regions. These regions are selected based on higher confidences from the prediction probability $W$, defined as the confidence score in Eq. (1), indicative of critical areas for the current cell type identified from the decoder in Eq. (2).

$$W = f(X; \theta)[:, 1] \qquad (1)$$

$$\text{top-}k(k, E, W, Y) = \{(e_1, w_1), (e_2, w_2), \ldots, (e_k, w_k)\} \cap Y \subseteq (E, W) \qquad (2)$$

In this model, $k$ represents the number of selected embedding features, $E$ denotes the embedding map from the last layer of the decoder, and $Y$ is the lay annotation. Each element in this map, $e$, represents the embedded features of a specific area in the image. $w$ represents the confidence scores associated with each embedding e.

Next, we calculate a cosine similarity score $S$ between the embedding from an arbitrary pixel and the embedding from critical embedding features, shown in Eq. (3). Here, $m$ denotes the channel of the feature embeddings.

$$S(e_i, e_{top-k}) = \frac{\sum_{m=1}^{M}(e_i \times e_{top-k})}{\sqrt{\sum_{m=1}^{M}(e_i)^2} \times \sqrt{\sum_{m=1}^{M}(e_{top-k})^2}} \qquad (3)$$

Given that labels from lay annotators can be noisy and erroneous, $W$ and $S$ in Eq. (4) are utilized in the subsequent equations to highlight the regions where both the model's predictions and lay annotations concur on the cell type, improving the accuracy of the segmentation. For expert annotations, higher confidence W and similarity S values lead to stronger weighting, maximizing precision. For lay annotations, where variability is higher, W and S help to mitigate noise by selectively emphasizing areas with agreement, thereby improving the robustness of the segmentation pipeline. This weighting approach is then integrated into the calculation of the loss function in Eq. (5):

$$\omega(W) = \exp(W) \times Y, \quad \omega(S) = S \times Y \tag{4}$$

$$L(Y, f(X; \theta)) = (L_{\text{Dice}}(Y, f(X; \theta)) + L_{\text{BCE}}(Y, f(X; \theta))) \times \omega(W) \times \omega(S) \tag{5}$$

## 4 Data and Experimental Design

We employed both public and in-house datasets for our experiments:

### 4.1 In-house Dataset

For this experiment, we utilized PAS-stained glomerular images paired with corresponding WT1 (podocyte marker) or GATA3 (mesangial cell marker) stained immunofluorescence (IF) images to enhance analytical accuracy. While IF imaging is somewhat more expensive than PAS staining, they are both affordable. It is important to note that once the model is trained, IF images are no longer required for new patients. The trained model only relies on PAS images as input, significantly reducing the ongoing costs. The dataset comprised 11 whole slide images (WSIs), including three slides of injured glomeruli. Digital scans of these stained tissue samples were conducted at 20× magnification. Using a comprehensive multi-modality multi-scale registration process, we created a dataset consisting of 1,147 patches featuring podocytes and 789 patches with mesangial cells. Each patch measured 512×512 pixels, derived from specific glomeruli or molecular structures within the WSIs.

WT1 and GATA3 exhibit heterogeneous expression in normal and diseased states. To mitigate the inconsistency, experienced pathologists provided us with the optimal threshold for different markers at the WSI level. Moreover, the introduction of the SAM method alleviates the inter-rater variabilities when applied to different stains.

Annotations were provided by one experienced renal pathologist and three computer science students, utilizing ImageJ (version v1.53t). The "Synchronize Windows" feature was employed for cursor synchronization across different modalities, and the "ROI Manager" was used to manage binary masks of each cell type. The dataset was randomly split into training, validation, and testing subsets in a 6:1:3 ratio, ensuring a balanced representation of both injured and normal glomeruli.

### 4.2 Public Dataset

For a more focused and detailed experiment on All-in-SAM, we utilized the MICCAI 2018 Monuseg dataset,[17] which includes 30 training images and 14 testing images. Each image has dimensions of 1000×1000 pixels and is accompanied by corresponding masks of nuclei.

The MICCAI dataset has inherent limitations in terms of its size, which may restrict the generalizability of our findings. We explicitly acknowledge this as a constraint and highlight the necessity of validating our results using larger and more diverse datasets in future studies. To address this limitation, we plan to expand our dataset by incorporating additional WSIs and annotated patches. Such efforts will facilitate more comprehensive evaluations and enhance the robustness of our findings in subsequent research.

*4.3 Environment and Evaluation Metrics*

The experiment involved delineating cellular structures on WSIs at a workstation equipped with a 12-core Intel Xeon W-2265 Processor and an NVIDIA RTXA6000 GPU. Separate annotation of cell contours was carried out using an 8-core AMD Ryzen 7 5800X Processor and an XP-PEN Artist 15.6 Pro Wacom tablet. The annotation process for a single cell type on one WSI was approximately 9 hours, while the batch processing of staining and scanning 24 IF WSIs took about 3 hours.

In this study, we use the Dice similarity coefficient (Dice) to evaluate pixel-level segmentation accuracy, and the F1-score is employed to assess instance-level detection.

## 5 Results

*5.1 All-in-SAM on In-house Multi-Class Cell Segmentation Dataset*

We present a comprehensive evaluation of the All-in-SAM method applied to multi-class cell segmentation. Shown in Figure 3 and Table 1 , we investigate the performance of four distinct methods, includes MOCL, SAM-L with tight bounding boxes and random bounding boxes, and All-in-SAM, applied to multi-class cell segmentation, specifically targeting podocytes and mesangials.

The F1-score represents the harmonic mean of precision and recall, two critical indicators of a model's accuracy. Each entry in the F1-score columns of the Table 1 reflects the F1-score calculated from the outcomes achieved by each segmentation method under specific conditions, such as "Injured Podocyte" and "Normal Mesangial."

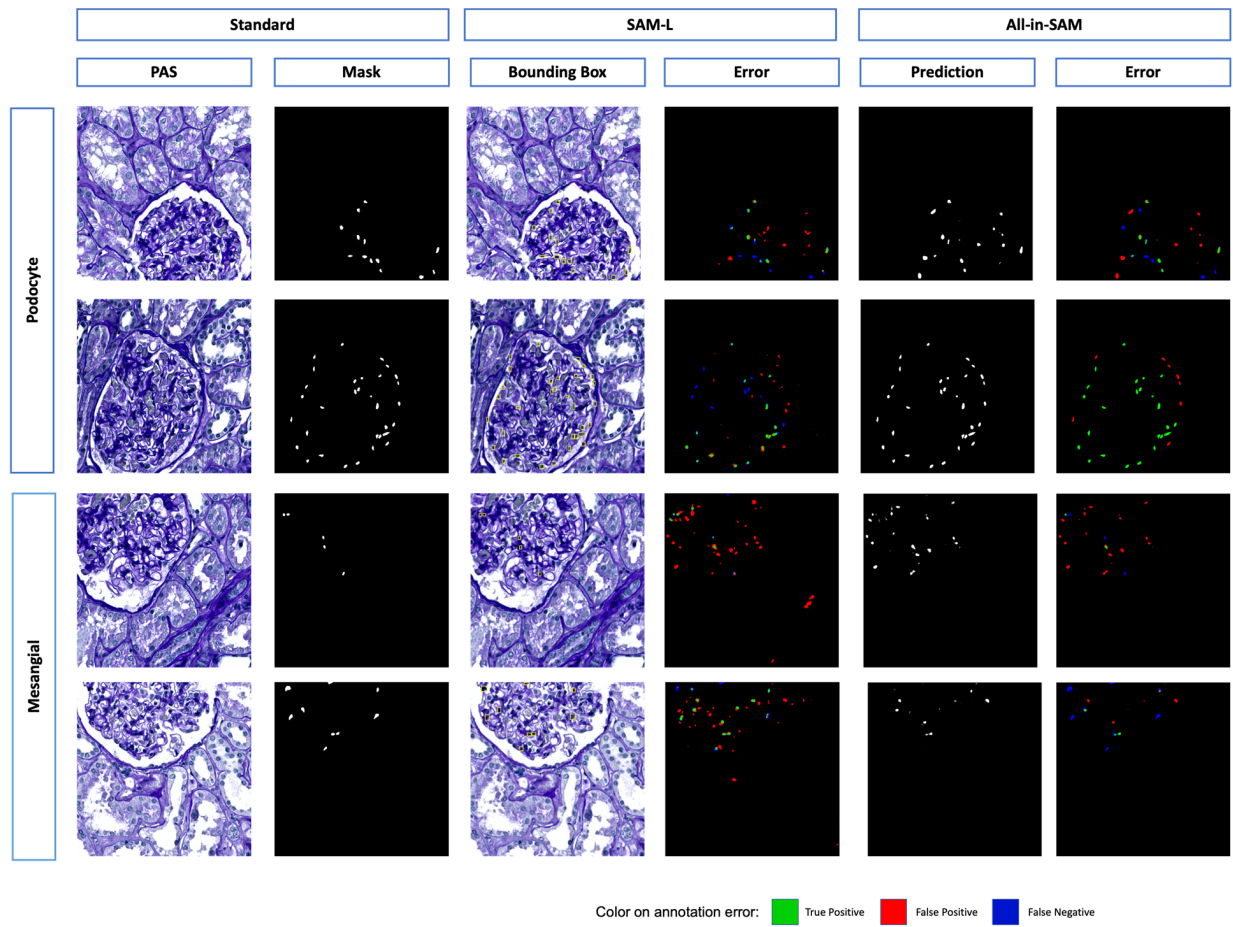

Fig 3: **Annotation results**: This figure presents the annotation accuracy using different methods, notably SAM-L and All-in-SAM, highlighting their effectiveness in various test conditions.

Table 1: Performance of Multi-Class Cell Segmentation (F1-score). The difference between the reference (Ref.) method and benchmarks is statistically evaluated by Wilcoxon signed-rank test.

|  |  | Pod. | Mes. | Pod. | Mes. | Pod. | Pod. | Mes. | Mes. |
|---|---|---|---|---|---|---|---|---|---|
| MOCL | Expert | 0.7124 | 0.7022 | 0.7547 | 0.6674 | 0.7321 | $p<0.05$ | 0.6685 | $p<0.05$ |
|  | Students | 0.7198 | 0.7157 | 0.7657 | 0.6830 | 0.7411 | $p<0.05$ | 0.7028 | $p<0.05$ |
| SAM-L tight | Expert | 0.7105 | 0.7027 | 0.7657 | 0.6683 | 0.7362 | $p<0.05$ | 0.6891 | $p<0.05$ |
|  | Students | 0.7043 | 0.7014 | 0.7390 | 0.6513 | 0.7205 | $p<0.05$ | 0.6817 | $p<0.05$ |
| SAM-L random | Expert | 0.7226 | 0.6994 | 0.7673 | 0.6713 | 0.7434 | $p<0.05$ | 0.6883 | $p<0.05$ |
|  | Students | 0.7170 | 0.7096 | 0.7565 | 0.6723 | 0.7354 | $p<0.05$ | 0.6949 | $p<0.05$ |
| All-in-SAM | Expert | 0.7218 | 0.7087 | 0.7699 | 0.7380 | **0.7462** | **Ref.** | **0.7146** | **Ref.** |
|  | Students | 0.7364 | 0.6987 | 0.7421 | 0.7187 | 0.7329 | $p<0.05$ | 0.7137 | $p<0.05$ |

We conducted a statistical analysis using the Wilcoxon signed-rank test to evaluate the performance differences in multi-class cell segmentation F1-scores. Specifically, we used the average podocyte F1-score derived from expert annotations in the All-in-SAM method as the reference value (0.7462). Similarly, the average mesangial F1-score derived from expert annotations in the All-in-SAM method (0.7146) was also used as a reference for mesangial comparisons. For each case, we calculated the differences between these reference values and the average F1-scores obtained from other methods.

The statistical test revealed that all comparisons for both podocytes and mesangials yielded p-values less than 0.05, indicating statistically significant differences between the All-in-SAM method and the other benchmarks. With the highest average F1-scores for both podocyte and mesangial cell segmentation based on expert annotations, All-in-SAM demonstrates superior precision and recall, consistently outperforming other methods across various conditions and annotator groups.

## 5.2 All-in-SAM on Public Dataset

Building on the insights from the initial results, we extend our analysis to compare All-in-SAM

with other leading methods under varying training regimes and annotation strengths. This comparison aims to assess the robustness and adaptability of All-in-SAM in less controlled environments.

Table 2 presents a further comparative analysis involving other SOTA methods such as nnUNet, LViT, and BEDs. The evaluation spans multiple training volumes and annotation qualities, from weak to complete, providing a comprehensive overview of each method's performance under diverse operational conditions."

Table 2: Comparison with other SOTA methods when using different numbers of training data with weak or complete annotation.

| Label | Method | Training Data | Dice | IoU | ADJ |
|---|---|---|---|---|---|
| Complete | Xie | All | - | - | 0.7063 |
| | | 30% | - | - | 0.6031 |
| | | 10% | - | - | 0.5501 |
| | LViT | All | 0.8033 | 0.6724 | - |
| | | 25% | 0.7994 | 0.6680 | - |
| | BEDs | All+More | 0.8177 | - | - |
| | nnUNet | All | 0.8244 | 0.6976 | 0.7028 |
| | | 4% | 0.7920 | 0.6540 | 0.6580 |
| | All-in-SAM | All | 0.8254 | 0.6974 | 0.7036 |
| | | 4% | 0.8134 | 0.6810 | 0.6867 |
| Weak | nnUNet | All | 0.8212 | 0.6935 | 0.6974 |
| | | 4% | 0.7913 | 0.6527 | 0.6562 |
| | All-in-SAM | All | 0.8246 | 0.6973 | 0.7024 |
| | | 4% | 0.8099 | 0.6760 | 0.6814 |

When trained with the entire dataset and complete annotation, both nnUNet and All-in-SAM exhibit comparable performance. However, with 100% of the training data volume, All-in-SAM emerges as the top performer. Subsequently, when the training data is reduced to 4%, All-in-SAM demonstrates superior performance compared to other methods. Notably, when utilizing weak labels for training data, All-in-SAM consistently outperforms other methods, maintaining the highest level of performance across evaluations.

5.3 Ablation studies

In Table 3, we specifically select two outstanding methods identified from Table 2: nnUNet and All-in-SAM, assessing their performance using both complete and weak labels across varying percentages of training data. And at this time, we use more metrics, like Recall and Precision to evaluate the model performance.

Table 3: Comparison of different methods trained by different numbers of weakly/completely annotated data.

| Label | Method | Training Data | Dice | AUC | Recall | Precision | bestF1 | IoU | ADJ |
|---|---|---|---|---|---|---|---|---|---|
| Complete | nnUNet | All | 0.8244 | 0.9604 | 0.8282 | 0.8255 | 0.8321 | 0.6976 | 0.7028 |
| | | 4% | 0.7920 | 0.9410 | 0.8490 | 0.7460 | 0.7970 | 0.6540 | 0.6580 |
| | | 0.50% | 0.7623 | 0.9249 | 0.8679 | 0.6861 | 0.7797 | 0.6135 | 0.6186 |
| | All-in-SAM | All | 0.8254 | 0.9717 | 0.8474 | 0.8081 | 0.8304 | 0.6974 | 0.7036 |
| | | 4% | 0.8134 | 0.9550 | 0.8492 | 0.7853 | 0.8190 | 0.6810 | 0.6867 |
| | | 0.50% | 0.7816 | 0.9440 | 0.8457 | 0.7351 | 0.7917 | 0.6379 | 0.6430 |
| Weak | nnUNet | All | 0.8212 | 0.9565 | 0.9126 | 0.7498 | 0.8368 | 0.6935 | 0.6974 |
| | | 4% | 0.7913 | 0.9474 | 0.9276 | 0.6922 | 0.8154 | 0.6527 | 0.6562 |
| | | 0.50% | 0.7500 | 0.9264 | 0.9336 | 0.6199 | 0.7757 | 0.5881 | 0.5918 |
| | All-in-SAM | All | 0.8246 | 0.9732 | 0.8947 | 0.7678 | 0.8339 | 0.6973 | 0.7024 |
| | | 4% | 0.8099 | 0.9522 | 0.8845 | 0.7509 | 0.8187 | 0.6760 | 0.6814 |
| | | 0.50% | 0.7873 | 0.9430 | 0.8704 | 0.7248 | 0.7982 | 0.6457 | 0.6502 |

When trained with complete training data, All-in-SAM demonstrates superior performance compared to nnUNet. This trend persists even when the training data is reduced to 0.5%. Additionally, when comparing All-in-SAM's performance between complete and weak labels, it maintains its effectiveness without any significant drop, indicating its robustness across different annotation qualities.

## 6 Discussion

The comprehensive empirical analysis conducted as part of this study highlights the profound efficacy of the molecular-empowered All-in-SAM Model in addressing the challenging task of multi- class cell segmentation within high-resolution WSIs. Our investigation spans a variety of testing scenarios characterized by different levels of annotation completeness and volumes of training data. The data distilled into Tables 2, 3, and 1 illustrate a clear superiority of our proposed models over current state-of-the-art (SOTA) methods, particularly emphasizing the robustness of our approaches when working with suboptimal annotations. Moreover, the proposed All-in-SAM model maintains high performance under constrained conditions, highlighting its adaptability and readiness for broader clinical adoption, which could significantly impact the future of digital pathology by making high-quality diagnostics more accessible and reliable.

The significance of this research lies not only in its technical contributions but also in its potential to transform pathological workflows. The ability of the molecular-empowered All-in-SAM model to reliably and efficiently segment WSIs addresses a critical bottleneck in digital pathology. By automating this traditionally labor-intensive process, the model enables faster, more accurate pathological assessments, which are vital for timely and effective medical diagnoses.

Moreover, this innovation holds particular promise for resource-limited settings where access to expert pathologists and advanced diagnostic tools may be scarce. By enhancing the accessibility and precision of digital pathology, the All-in-SAM model can democratize high-quality diagnostics, bridging gaps in healthcare inequities and improving outcomes on a global scale. Its ability to operate effectively with limited training data and annotations further amplifies its utility in settings with restricted resources, accelerating the pace of diagnostic processes while maintaining reliability.

In conclusion, the molecular-empowered All-in-SAM model represents a significant step forward in digital pathology. Its robust performance, adaptability, and accessibility underscore its potential to advance medical diagnostics, streamline workflows, and ultimately contribute to improving healthcare outcomes worldwide. Future work will focus on expanding the dataset and validating the model across diverse clinical and pathological contexts to further enhance its generalizability and impact.

## 7 Conclusions

In this study, we have introduced the molecular-empowered All-in-SAM Model, a novel framework designed to address the challenges of multi-class cell segmentation within high-resolution WSIs in computational pathology. By leveraging innovative techniques such as the SAM and MOCL, the key contributions lie in its ability to utilize lay annotators for cost-effective data collection with only weak labels and molecular-guided segmentation, employ SAM adaptor for label-efficient finetuning, and apply advanced corrective learning techniques for precise segmentation tasks. Through comprehensive evaluations, the proposed All-in-SAM Model has demonstrated superior performance while reducing reliance on specialized annotations from domain experts.

**Disclosures**

The authors of the paper have no conflicts of interest to report.

**Code and Data Availability**

The code used to generate the results and figures is available in a Github repository (https://github.com/Xueyuan33/Molecular-Empowered-All-in-SAM). We have 2 different datasets: In-house Dataset and Public Dataset. The In-house Dataset that support the findings of this article are not publicly available due to privacy. They can be requested from the author at lixueyuan69@gmail.com. For the Public Dataset, you can find it from MICCAI 2018 Monuseg dataset.

**Acknowledgments**

This research was supported by NIH R01DK135597(Huo), DoD HT9425-23-1-0003(HCY). This work was also supported by Vanderbilt Seed Success Grant, Vanderbilt Discovery Grant, and VISE Seed Grant. This project was supported by The Leona M. and Harry B. Helmsley Charitable Trust grant G-1903-03793 and G-2103-05128. This research was also supported by NIH grants R01EB033385, R01DK132338, REB017230, R01MH125931, and NSF 2040462. We extend gratitude to NVIDIA for their support by means of the NVIDIA hardware grant. This works was also supported by NSF NAIRR Pilot Award NAIRR240055.

**Biography**

**Xueyuan Li** is currently a master student in Data Science and at Vanderbilt University. She is advised by Prof. Yuankai Huo at HRLB Lab. Her main research interests are medical image analysis, deep learning, and computer vision.

Biographies and photographs of the other authors are not available.

# List of Figures

## List of Tables